\documentclass{article}

\usepackage{spconf}
\usepackage{subimages}
\usepackage{graphicx}
\setfigdir{figs}

\usepackage[cmex10]{amsmath}

\usepackage{amsthm}

\usepackage{array}
\usepackage{booktabs}
\usepackage{multirow}

\usepackage{calligra}
\DeclareMathAlphabet{\mathcalligra}{T1}{calligra}{m}{n}

\usepackage{color}

\usepackage{amssymb}

\usepackage{fancyhdr}

\title{FAST-FORWARD VIDEO BASED ON SEMANTIC EXTRACTION}
\name{Washington L.S. Ramos*, Michel M. Silva*, Mario F. M. Campos, Erickson R. Nascimento \thanks{*First two authors contributed equally.}}
\address{Universidade Federal de Minas Gerais (UFMG), Brazil 
	\\ \{washington.ramos, michelms, mario, erickson\}@dcc.ufmg.br}
\begin{document}
%
\maketitle

\thispagestyle{fancy}
\fancyhf{}
\chead{In Proceedings of the 2016 IEEE International Conference on Image Processing (ICIP) \\ The final publication is available at: http://dx.doi.org/10.1109/ICIP.2016.7532977}
\setlength{\headsep}{0.08 in}

\begin{abstract}

Thanks to the low operational cost and large storage capacity of smartphones and wearable devices, people are recording many hours of daily activities, sport actions and home videos. These videos, also known as egocentric videos, are generally long-running streams with unedited content, which make them boring and visually unpalatable, bringing up the challenge to make egocentric videos more appealing. In this work we propose a novel methodology to compose the new fast-forward video by selecting frames based on semantic information extracted from images. The experiments show that our approach outperforms the state-of-the-art as far as semantic information is concerned and that it is also able to produce videos that are more pleasant to be watched.

\end{abstract}

\section{Introduction}
\label{sec:introduction}

Thanks to advances in technology which constantly leads to the decreasing cost of wearable and mobile cameras and the increase in storage capacity, first person videos have become much more ubiquitous in social media, video-sharing websites and personal repositories. Wearable devices such as GoPro\texttrademark\  cameras and Google Glass\texttrademark\  can be operated with no intervention which opens up unprecedented ways for users to record many continuous hours of daily activities (e.g. walking, driving), sport actions (e.g. climbing, running, bicycling), home videos (e.g. weddings, family meetings, birthdays) and monitoring tasks (e.g. police patrol and life guard). These videos are referred to as egocentric videos.

Usually most egocentric videos are exhibited with no post-processing or editing which makes them hard and boring to watch since they are long, monotonous and subject to camera instability~\cite{pol_hal_aro_pel}. Over the last couple of years the development of methods to speed-up egocentric videos has become a research topic since the use of simple fast-forward methods such as frame sub-sampling  at a fixed rate produces jerky videos which are at best, visually unpleasant to watch. 

Several works have been recently proposed to tackle the instability and to create watchable egocentric videos by automatically selecting frames to compose a more compact final video. The challenge faced by the proposed methodologies is that not all frames of the video contains information that are equally relevant. For example, a camera installed on a police car would be recording all day long but with only few events of interest such as when the officer interacts with someone or engages in police activity (e.g. pursuit and capture). Virtually all \textit{hyperlapse} algorithms do not select frames according to their relevance to the viewer, but instead treat each all frames as equally important.

The contribution of this work is a novel methodology capable of transforming raw egocentric videos into watchable videos by considering both the pleasantness and relevance of frames to the viewer. Our approach analyses the semantic information extracted from the frames and segments the video by selecting the set of images which maximizes the semantic term, the required speed-up as well as the desired smoothness in the transition between the frames.

\paragraph*{Related Work.}
\label{sec:related_work}

In the past several years, video summarization methods have been the main technique used to create a short output video from a long input one with the goal of maintaining essential information while saving the viewer a considerable amount of viewing time~\cite{lee_gho_gra, zha_roy, mei_gua_wan_wan}. Although these video summarization methods have been increasing their ability to select relevant frames to represent the whole video, the final result is, in general, a set of discontinuous frames.

Recent efforts to create smooth fast-forward videos from egocentric videos can be divided into two main categories: reconstruction of a 3D model of the scene along with the creation a smooth path with a virtual camera and adaptive selection of a frame set that generates a smoother final video.

A representative method of the former category is the work of Kopf et al.~\cite{kop_coh_sze}. The authors present a technique based on estimating a 3D model of the scene and generating camera poses to optimize a new and smoother path followed by a virtual camera. Although the final video is very smooth, the method creates many artifacts due to the large number of interpolated areas in the virtual camera's path. The technique also requires camera motion and parallax to compute the 3D model of the scene.

The Instagram Hypelapse App~\cite{kar_}, the work of Joshi et al.~\cite{jos_kie_toe_mik_mat_coh} and the work of Poleg et al.~\cite{pol_hal_aro_pel} are recent examples of the selection approaches category. The Instagram Hypelapse App combines video stabilization~\cite{kar_jac_bae_lev} and the phone gyroscope to produce \textit{hyperlapse} videos. The major limitation of this approach is the need of inertial data, which makes it unfeasible to be used in videos recorded using a general camera. Joshi et al. create a \textit{hyperlapse} using an optimization function. To do so, they calculate a frame matching using sparse feature-based techniques to verify the alignment of one frame with the others. Then they apply a dynamic-time-warping algorithm to find an optimal smooth path. The Poleg et al.'s approach creates a graph from the original video and computes the shortest path in order to find the best frames to compose the \textit{hyperlapse}. The frames of the video are taken as the nodes of the graph and edges are the relation between frames. 

Although the aforementioned solutions succeed in speeding up long videos and producing a result that is pleasant to watch, they do not take into account the fact that some frames are more important than others, which is related to the semantic in regions of the scene. This work presents a new method that performs the frame selection based on semantic information without degenerating the smoothness of the video.

\section{Methodology}
\label{sec:methodology}

\begin{figure*}[t!]
	\centering
	\includegraphics[width=\textwidth]{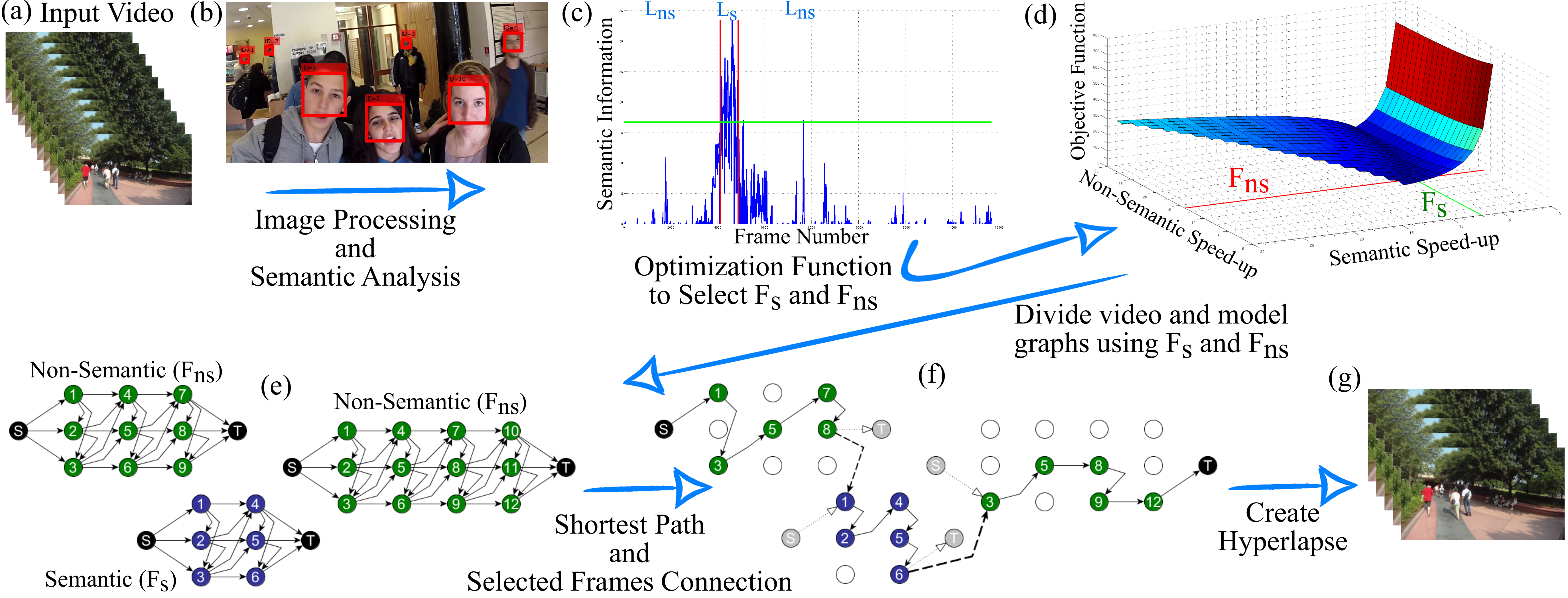}
	\caption{Overview of our fast-forward methodology. From the input video (a) ROIs containing the semantic information are detected in each frame and semantic scores are computed (b). Theses scores are used to segment the video into semantic and non-semantic parts (c). Based on the length of each part, speed-up rates are estimated, one for semantic and the other for the non-semantic part (d). Each part produces a graph from the frames (e) where the shortest path algorithm is applied (f) to create the final video (g).}
	\label{fig:methodology}
	\vspace{-1.0em} 
\end{figure*}

Our methodology is illustrated in Fig.~\ref{fig:methodology}. From each frame of the input video, we extract the semantic information, which is encoded by the score function $\mathcal{S}: \mathbb{N} \rightarrow \mathbb{R}$. This function is composed of three components: i) the confidence of the extracted information; ii) the centrality of the analyzed region, since the input is an egocentric video, the central area of the images should have a higher weight and; iii) the size of the region, since larger areas means that the subject of interest is closer to the recorder, therefore its score is higher. The final semantic score for the frame $i$ and $K$ detected Regions of Interest (ROI) (where the semantic information will be extracted) is given by:
\begin{equation}
	\mathcal{S}_{i} = \sum_{k \in f_{i}} C(k) \cdot G_{\sigma}(k) \cdot A(k),
	\label{eq:frame_semantic_score}
\end{equation}
\noindent where $C(k)$ returns the confidence of the $k$-th ROI at the frame $f_{i}$, $G_{\sigma}(k)$ is the value of the Gaussian function centered in the frame $f_{i}$ at the central position of the $k$-th ROI and $A(k)$ computes the area of the $k$-th ROI around the semantic information in the frame $f_{i}$. An example using a face detector is depicted in the Fig.~\ref{fig:methodology}-b. 

Then, we split the video into semantic and non-semantic segments (vertical red lines in Fig.~\ref{fig:methodology}-c). After filtering by using a Gaussian function to remove possible outliers, the mean value between the minimum and maximum peaks is used to define a threshold (horizontal green line). The video segments above this threshold are defined as semantic parts and the remaining segments are classified as non-semantic parts. To keep the semantic segments smoother, we use two speed-up goals: the $\mathcal{F}_{s}$ values the segments classified as semantic and $\mathcal{F}_{ns}$ values those classified as non-semantic. These values are computed by the minimization of the Equation~\ref{eq:DesiredSpeedUp}:
\begin{equation}
	D\left(\mathcal{F}_{ns}, \mathcal{F}_{s} \right) = \left(\left|\frac{L_{s}+L_{ns}}{F_{d}} - \frac{L_{s}}{\mathcal{F}_{s}} - \frac{L_{ns}}{\mathcal{F}_{ns}} \right|\right),
	\label{eq:DesiredSpeedUp}
\end{equation}
\noindent where $L_{s}$ is the length (in number of frames) of the semantic parts, $L_{ns}$ is the length of the non-semantic parts, $F_{d}$ is the desired speed-up for the whole video (defined by the user).

The minimum of the Equation~\ref{eq:DesiredSpeedUp} occurs when the number of frames of the whole video divided by the desired speed-up is equal to the sum of the semantic parts divided by the semantic speed-up with the non-semantic parts divided by the non-semantic speed-up. Since this function has many minimum points, we solve it according to our requirements through the following function:
\begin{equation}
	\underset{\mathcal{F}_{s},\mathcal{F}_{ns}}{\arg\min} \left(D\left(\mathcal{F}_{ns}, \mathcal{F}_{s}\right) + \lambda_{1} \cdot |\mathcal{F}_{ns}-\mathcal{F}_{s}| + \lambda_{2} \cdot |\mathcal{F}_{s}|\right),
	\label{eq:argmin}
\end{equation}
\noindent where the second term in the objective function restricts the value of $\mathcal{F}_{s}$ to be not too different from $\mathcal{F}_{ns}$, the third term restricts the $\mathcal{F}_{s}$ to be as small as possible and, $\lambda_1$ and $\lambda_2$ are regularization terms.

The Fig.~\ref{fig:methodology}-d shows an example of the search space for the objective function. To solve the optimization problem we add some space restrictions: (1) $\mathcal{F}_{s} \leqslant F_{d}$, once we want a lower speed-up ratio in the semantic parts; (2) $\mathcal{F}_{s} \leqslant \mathcal{F}_{ns}$, because we want to sample at a higher ratio in the semantic parts rather than the opposite; (3) once we want to keep the $F_{d}$ for the whole video $\mathcal{F}_{ns} \geqslant \mathcal{F}_{d}$. With these restrictions, the problem becomes easier to be solved, because the search space is discrete and finite.

We model each video segment using a weighted graph similar to Poleg et al.. Each frame represents a node in the graph and the temporal relation between two nodes $i$ and $j$ are represented by an edge. The edges are created up to a temporal distance $\tau_{max}$ and represent the cost of including the frame $j$ after the frame $i$ in the final video. The weight of an edge is determined by: 1) the balance term $\mathcal{B}_{i,j}$, that indicates the average distance of the focus of expansion (FOE) from the center of the image; 2) the velocity cost term $\mathcal{V}_{i,j}$, that indicates the speed of motion from frame $i$ to $j$, where the desired speed is given by the magnitude of the optical flow between $i$ and $j$; 3) the appearance cost term $\mathcal{A}_{i,j}$, that measures the similarity between two frames (the Earth Mover's Distance is used as the similarity function); 4) the semantic cost term $\mathcal{S}_{i,j}$ that computes the semantic relation between frames $i$ and $j$ and; 5) this term penalizes skips greater than the desired speed-up. 

The first three terms are computed similar to Poleg et al. and the semantic term is given by the Equation~\ref{eq:semantic_term}:
\begin{equation}
	\mathcal{S}_{i,j} = \frac{1}{\mathcal{S}_{i} + \mathcal{S}_{j} + \epsilon},
	\label{eq:semantic_term}
\end{equation}
\noindent where $\mathcal{S}_k$ is the semantic score of frame $k$ and $\epsilon$ is used to prevent division by zero for non-semantic frames. 

The semantic term increases whenever frame $i$ or $j$ has a small cost. An example of the modeled graph can be seen in the Fig.~\ref{fig:methodology}-e. The final weight of edges $i$ to $j$ is given by:
\begin{equation}
	\mathcal{W}_{i,j} = (\alpha \cdot \mathcal{B}_{i,j} + \beta \cdot \mathcal{V}_{i,j} + \gamma \cdot \mathcal{A}_{i,j} + \eta \cdot \mathcal{S}_{i,j}) \cdot \left\lceil{\frac{(j-i)}{\mathcal{F}}}\right\rceil,
	\label{eq:graph_formulation}
\end{equation}
\noindent where $\alpha$, $\beta$, $\gamma$, $\eta$ are the weights of the cost terms and $\mathcal{F}$ indicates the estimated speed-up for that graph.

The minimization in the whole graph is computed by running a shortest path algorithm (in this work we used Bellman-Ford) in each segment (Fig.~\ref{fig:methodology}-f). The selected frames of each segment compose the final video (Fig.~\ref{fig:methodology}-g).

\section{Experiments}
\label{sec:experimental_results}

In this section we present the results of our methodology and compare it against prime works in the literature running on $9$ publicly available video sequences: Bike 1, Bike 2, Bike 3, Walking 1 and Walking 2~\cite{kop_coh_sze}; Running, Driving and Walking 3~\cite{pol_hal_aro_pel} and; Walking 4~\cite{pol_aro_pel}.

In our work, the semantic information was extracted from the region defined by the ROI of detected human faces. We used the state-of-the-art face detector Normalized Pixel Difference~\cite{lia_jai_li} due to its high accuracy in the wild and its low computational cost. We applied the confidence score given for each detected face as the $C(k)$ value. In order to reduce the false positive rate, we filtered the outputs of the face detector by removing all detections with score smaller than $\theta=10$. Faces in the interval $\{\theta=10, \zeta=60\}$ were kept if they were detected frequently in a short span of time. All detections with confidence greater than $\zeta$ were considered faces. The function used in Equation~\ref{eq:frame_semantic_score} is a Normal with mean zero and $\sigma=\max(W/2,H/2)$ where $W$ is the video width and $H$ is the video height. The $\tau_{max}$ value was set to $100$. To prevent division by zero, we set $\epsilon = 1$.

We compared our result (with a desired speed-up of $10$) against four different techniques: i) \textit{Na{\"i}ve}, which simply creates a video by taking every $10$-th frame of the input video. This selection gives us the perfect speed-up, but the jitter and the semantic content are video-dependant values; ii) \textit{Na{\"i}veFaces}, which creates a video by taking iteratively the frames with the higher semantic score until the desired speed-up is achieved. This technique gives us the best frame selection for semantic content and the desired speed-up, however, some parts of the video are completely removed, thus the final result does not represent the whole video; iii) \textit{EgoSampling}, which creates a video by using the Poleg et al.'s technique~\cite{pol_hal_aro_pel} with parameters defined according to the best values of their work; iv) \textit{Microsoft Hyperlapse} (MH)~\cite{jos_kie_toe_mik_mat_coh}, where we used their software to create the output videos.


We quantified the performance of the methods according to the following metrics:
\textit{(1) Semantic Amount}: the semantic information present in the final video is quantified by summing up the semantic score of every output frame;
\textit{(2) Jitter Amount}: it is measured between consecutive output frames through the mean magnitude of the FOE locations differentiation. The lower is the amount of jitter the smoother is the video;
\textit{(3) Speed-up Deviation}: it is the distance between the desired and the achieved speed-up of each technique.

\paragraph*{Results.} 

\begin{figure}[t!]
	\centering
	\includegraphics[width=0.43\textwidth]{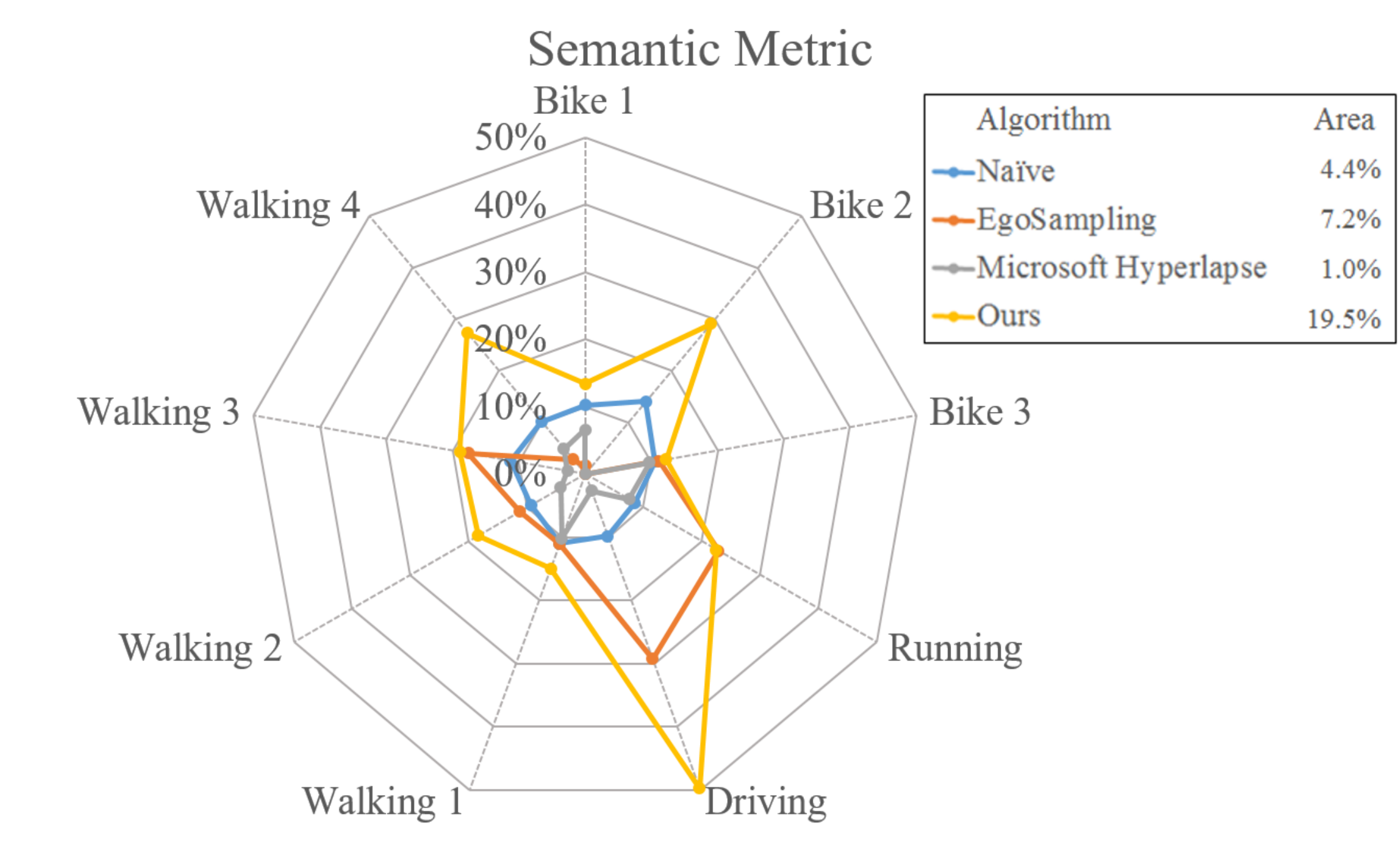}
	\caption{Semantic information in the final video. The values are related to Na{\"i}veFaces result, once it selects as much information as possible. }
	\vspace{-0.9em} 
	\label{fig:semantic}
\end{figure}

\begin{figure}[t!]
	\centering
	\includegraphics[width=0.5\textwidth]{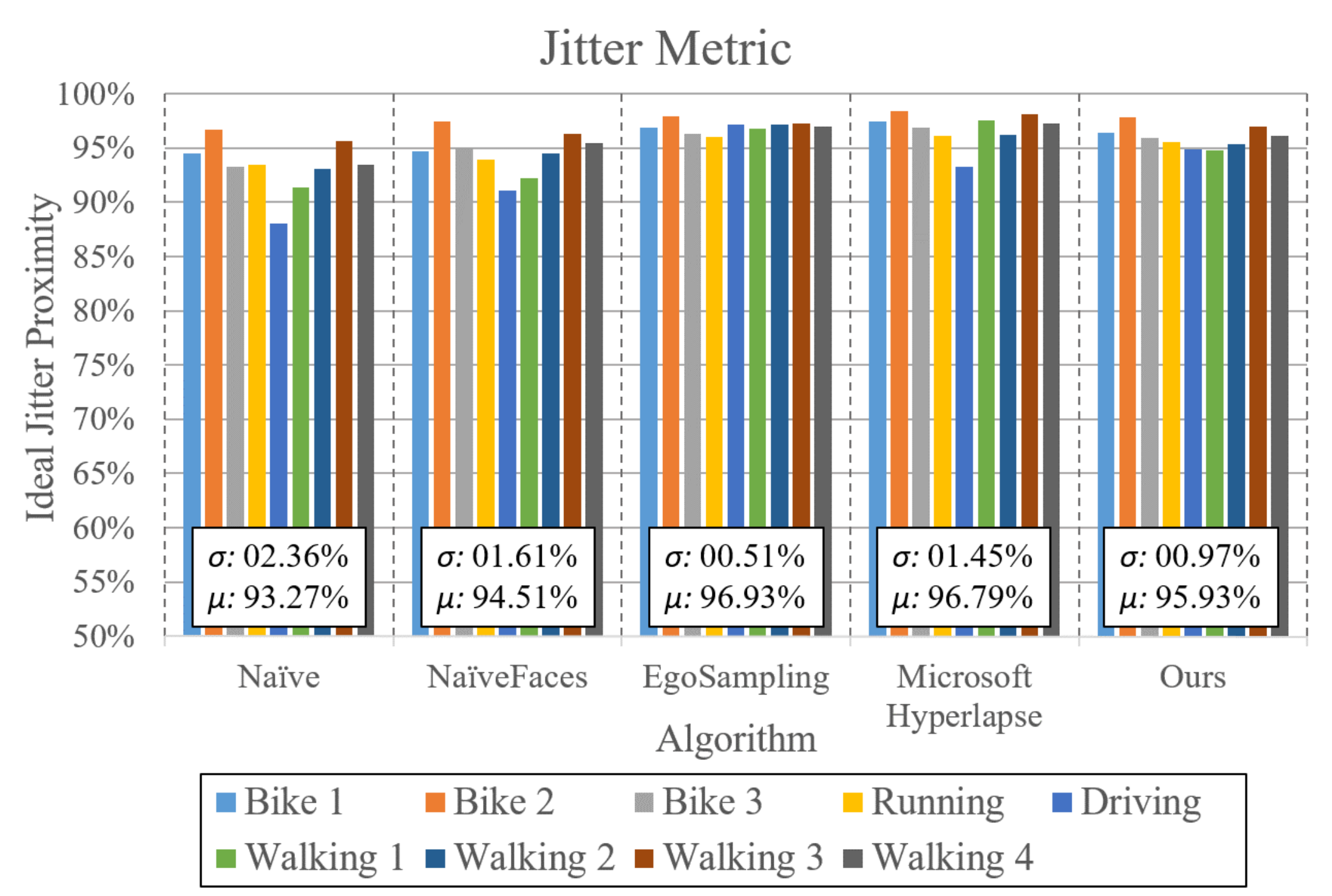}
	\caption{Jitter analysis over the sequences. As expected, the jitter of Na{\"i}ve and Na{\"i}veFaces were worse than the others. Our method is statistically tied with EgoSampling and MH.}
	\label{fig:jitter}
\end{figure}

Fig.~\ref{fig:semantic} depicts the results for the semantic metric, which are presented with relation to the score achieved by Na{\"i}veFaces, since it leads to frame sequence with the highest semantic information possible for the desired speed-up. Our result is better than the others, achieving an improvement around $27$ percentage points. In one sequence we managed to keep $50\%$ of the semantic information selected by Na{\"i}veFaces, and moreover, we got a closer approximation of the ideal jitter (see Fig. \ref{fig:jitter}), which is not achieved by Na{\"i}veFaces. Our method area is almost 3 times larger than the area of the second best ranked method.

We calculated the jitter for a hypothetic output video where for every consecutive frames the FOE location is as far as possible one from another and we used this value as the worse jitter possible. Thus, the percentage of improvement over this value is used to show our results, which are presented in the Fig.~\ref{fig:jitter}. Our output videos are as smooth as the EgoSampling and MH outputs are for most of the sequences, differently from Na{\"i}ve and Na{\"i}veFaces that achieve poor results in this measure. In some sequences, Na{\"i}veFaces achieves a good result because of its solid selection of frames in some segments of video. MH creates the smoothest sequences in the most of the cases, but the differences between their method and ours are not huge, being around 2 percentage points. The same happens when comparing EgoSmapling to ours, since we emphasize the semantic.

\begin{figure}[t!]
	\centering
	\includegraphics[width=0.47\textwidth]{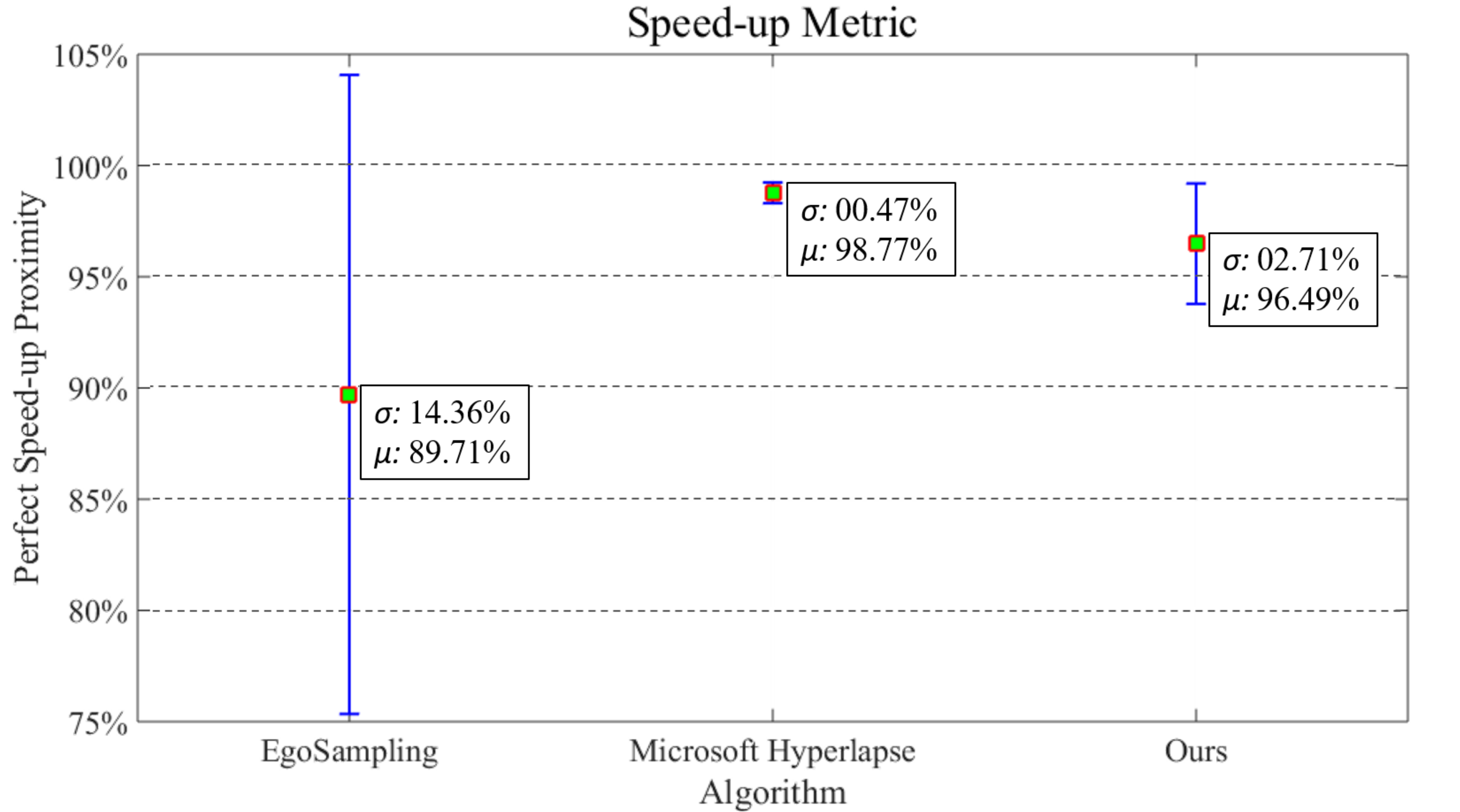}
	\caption{Mean and standard deviation of the speed-up measured over all sequences for each method. The values show how close the algorithms are from the ideal. }
	\vspace{-1.3em} 
	\label{fig:speedupAlt}
\end{figure}

Because $\tau_{max}$ represents the maximum allowed skip of the frames, it is the maximum speed-up that could be achieved. The worst speed-up deviation is given by the absolute difference between the desired speed-up and $\tau_{max}$. Fig.~\ref{fig:speedupAlt} presents the mean and standard deviation values for each technique, where the percentage indicates how good is the deviation in comparison to the worst deviation possible. For a better visualization, we removed the perfect speed-up deviation techniques Na{\"i}ve and Na{\"i}veFaces.

In general, our output videos present a slightly small difference when compared with the MH, which is the best one in this metric. EgoSampling has a low mean and a high standard deviation, which means that it is inaccurate in this metric.

\section{Conclusions}
\label{sec:conclusions_and_future_works}
We presented a novel method for producing \textit{hyperlapse} videos from egocentric videos focusing on its semantic content. Our method analyzes the semantic score of each frame and segments the video into semantic and non-semantic parts. Based on the length of each segment and the desired speed-up for the final video, we solve an optimization function to select different speed-ups for each type of segment. The frame selection is performed by a shortest path algorithm. As far as semantic metric is concerned, the final video created by our methodology was superior.

\section{Acknowledgments}
The authors would like to thank the agencies CAPES, CNPq, FAPEMIG, ITV (Vale Institute of Technology) and Petrobras for funding different parts of this work. 

\bibliographystyle{IEEEbib}
\bibliography{ICIP_2016_Washington_Michel}

\begin{thebibliography}{10}

\bibitem{pol_hal_aro_pel}
Yair Poleg, Tavi Halperin, Chetan Arora, and Shmuel Peleg,
\newblock ``Egosampling: Fast-forward and stereo for egocentric videos,''
\newblock in {\em Computer Vision and Pattern Recognition (CVPR), 2015 IEEE
  Conference on}, June 2015, pp. 4768--4776.

\bibitem{lee_gho_gra}
Yong~Jae Lee, Joydeep Ghosh, and Kristen Grauman,
\newblock ``Discovering important people and objects for egocentric video
  summarization,''
\newblock in {\em Computer Vision and Pattern Recognition (CVPR), 2012 IEEE
  Conference on}, June 2012, pp. 1346--1353.

\bibitem{zha_roy}
Shu Zhang and Amit~K. Roy-Chowdhury,
\newblock ``Video summarization through change detection in a non-overlapping
  camera network,''
\newblock in {\em Image Processing (ICIP), 2015 IEEE International Conference
  on}, Sept 2015, pp. 3832--3836.

\bibitem{mei_gua_wan_wan}
Shaohui Mei, Genliang Guan, Zhiyong Wang, Shuai Wan, Mingyi He, and David~Dagan
  Feng,
\newblock ``Video summarization via minimum sparse reconstruction,''
\newblock {\em Pattern Recognition}, vol. 48, no. 2, pp. 522 -- 533, February
  2015.

\bibitem{kop_coh_sze}
Johannes Kopf, Michael~F. Cohen, and Richard Szeliski,
\newblock ``First-person hyper-lapse videos,''
\newblock {\em ACM Trans. Graph.}, vol. 33, no. 4, pp. 78:1--78:10, July 2014.

\bibitem{kar_}
A.~Karpenko,
\newblock ``The technology behind hyperlapse from instagram,''
  http://instagram-engineering.tumblr.com/post/95922900787/hyperlapse, Aug.
  2014,
\newblock Accessed: 2016-05-12.

\bibitem{jos_kie_toe_mik_mat_coh}
Neel Joshi, Wolf Kienzle, Mike Toelle, Matt Uyttendaele, and Michael~F. Cohen,
\newblock ``Real-time hyperlapse creation via optimal frame selection,''
\newblock {\em ACM Trans. Graph.}, vol. 34, no. 4, pp. 63:1--63:9, July 2015.

\bibitem{kar_jac_bae_lev}
A.~Karpenko, D.~Jacobs, J.~Baek, and M.~Levoy,
\newblock ``Digital video stabilization and rolling shutter correction using
  gyroscopes,''
\newblock in {\em Stanford University Computer Science Tech Report CSTR
  2011-03}, 2011.

\bibitem{pol_aro_pel}
Yair Poleg, Chetan Arora, and Shmuel Peleg,
\newblock ``Temporal segmentation of egocentric videos,''
\newblock in {\em Computer Vision and Pattern Recognition (CVPR), 2014 IEEE
  Conference on}, June 2014, pp. 2537--2544.

\bibitem{lia_jai_li}
Shengcai Liao, Anil~K. Jain, and Stan~Z. Li,
\newblock ``A fast and accurate unconstrained face detector,''
\newblock {\em Pattern Analysis and Machine Intelligence, IEEE Transactions
  on}, vol. 38, no. 2, pp. 211--223, Feb. 2016.

\end{thebibliography}

\end{document}